\documentclass[conference]{IEEEtran}
\IEEEoverridecommandlockouts
\usepackage{amssymb}
\usepackage{cite}
\usepackage{amsmath,amssymb,amsfonts}
\usepackage{algorithmic}
\usepackage{graphicx}
\usepackage{textcomp}
\usepackage{xcolor}
\usepackage{multirow}
\usepackage{booktabs}
\usepackage{diagbox}
\usepackage{array}
\usepackage{hyperref}
 
\hypersetup{
    colorlinks=true, linkcolor=black,
    urlcolor=black,
    citecolor=black,
}

\def\BibTeX{{\rm B\kern-.05em{\sc i\kern-.025em b}\kern-.08em
    T\kern-.1667em\lower.7ex\hbox{E}\kern-.125emX}}
\begin{document}

\title{MVLLaVA: An Intelligent Agent for Unified and Flexible Novel View Synthesis\\
}


\author{\IEEEauthorblockN{Hanyu Jiang, Jian Xue, Xing Lan, Guohong Hu and Ke Lu\thanks{\IEEEauthorrefmark{2}Corresponding Author}\IEEEauthorrefmark{2}}
\IEEEauthorblockA{\textit{School of Engineering Science,} 
\textit{University of Chinese Academy of Sciences,} 
Beijing, China \\
Email: \{jianghanyu231, lanxing19, huguohong22\}@mails.ucas.ac.cn, \{xuejian,  luk\}@ucas.ac.cn}
}

\maketitle

\begin{abstract}
This paper introduces MVLLaVA, an intelligent agent designed for novel view synthesis tasks. MVLLaVA integrates multiple multi-view diffusion models with a large multimodal model, LLaVA, enabling it to handle a wide range of tasks efficiently. MVLLaVA represents a versatile and unified platform that adapts to diverse input types, including a single image, a descriptive caption,  or a specific change in viewing azimuth, guided by language instructions for viewpoint generation. We carefully craft task-specific instruction templates, which are subsequently used to fine-tune LLaVA. As a result, MVLLaVA acquires the capability to generate novel view images based on user instructions, demonstrating its flexibility across diverse tasks.
Experiments are conducted to validate the effectiveness of MVLLaVA, demonstrating its robust performance and versatility in tackling diverse novel view synthesis challenges.

\end{abstract}


\section{Introduction}
Novel view synthesis plays a crucial role in computer vision, enabling the generation of images from new viewpoints. 
Recent methods like Neural Radiance Fields (NeRF) \cite{mildenhall2021nerf,muller2022instant,wang2021nerf,pumarola2021d} and 3D Gaussian splatting (3DGS)\cite{kerbl20233d,yu2024mip,li2024advances} focus on directly optimizing a 3D representation of a scene by reconstructing it from multiple 2D images. Although these methods produce high-quality 3D representations, they typically require 
many images for each scene and involve intensive optimization processes. 
In contrast, much research has shifted towards more efficient multi-view diffusion models \cite{watson2022novel,liu2023zero,shi2023zero123++,shi2023mvdream,wang2023imagedream,gao2024cat3d,di2024dimvis,tseng2023consistent} for novel view synthesis. 
\begin{figure}[t]
\centerline{\includegraphics[width=\linewidth]{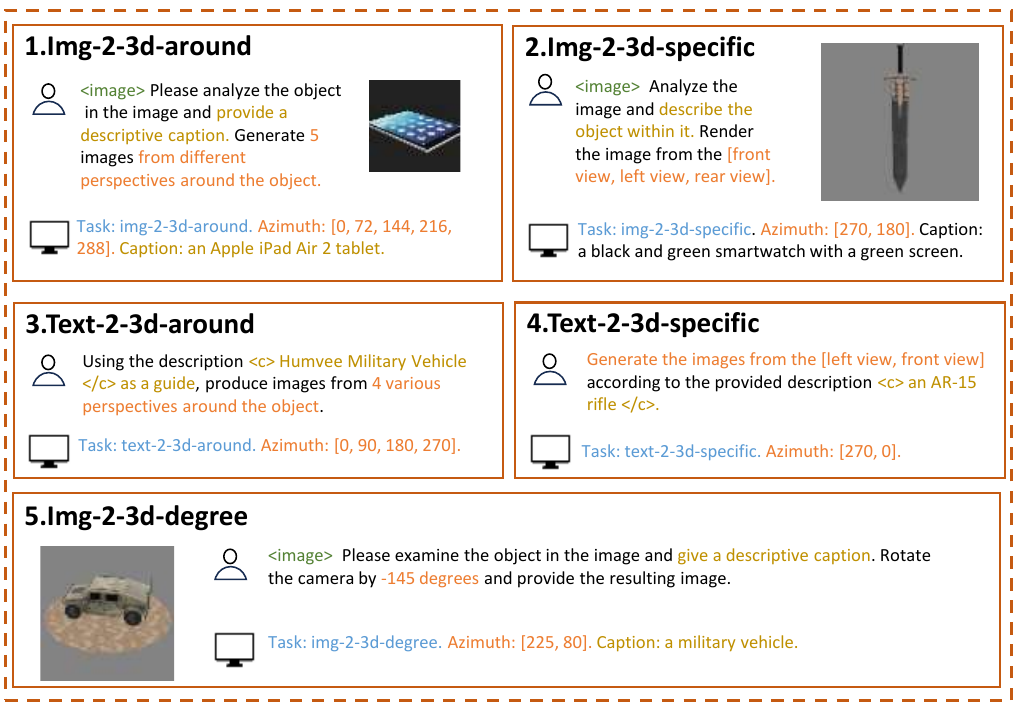}}
\caption{Application cases of MVLLaVA in novel view synthesis: MVLLaVA generates novel view images around an object or from specified viewpoints, using a reference image or caption. It can also
rotate the camera to produce the desired image. MVLLaVA unifies  multi-view generation capabilities to synthesize novel views, with user-friendly and intuitive instructions.}
\label{tasks}
\end{figure}

Despite their potential, current multi-view diffusion models for novel view synthesis still face limitations in task specificity and versatility. For instance, ImageDream\cite{wang2023imagedream} requires both an image and a caption to generate new views. When only a caption is provided, the process becomes cumbersome, as a separate T2I diffusion model\cite{rombach2022high} must first generate an image, which is then fed into ImageDream. This not only reduces workflow efficiency but also limits the model's ability to handle diverse inputs. Additionally, these models lack user-friendliness, offering limited control over viewpoint selection and insufficient support for natural language instructions, resulting in a less intuitive and flexible user experience.

To address these issues, we introduce MVLLaVA\footnote{Project page: \url{https://jamesjg.github.io/MVLLaVA_homepage/}}, an intelligent agent that seamlessly integrates multiple multi-view diffusion models with a large multimodal model LLaVA \cite{liu2024visual,liu2024improved}. 
MVLLaVA is designed to handle a range of task-specific instructions, which is achieved by developing view-based instruction templates and fine-tuning LLaVA to improve its adaptability and performance.
After instruction tuning, MVLLaVA intelligently interprets the input and selects the appropriate downstream model to generate the corresponding multi-view images. 
This task-driven approach, as shown in Fig. \ref{tasks}, enhances the flexibility and scalability of the generation process. It enables MVLLaVA to seamlessly adapt to diverse inputs and user requirements, effectively addressing the limitations of existing multi-view diffusion models.



In summary, the contributions of this paper are as follows:

\begin{itemize} 

\item An intelligent agent MVLLaVA is proposed for novel view synthesis, which integrates multiple task-specific diffusion models with LLaVA.

\item The instruction templates tailored to different tasks are designed, enabling MVLLaVA to handle a range of diverse and flexible instructions with robustness.
    
\item The effectiveness of MVLLaVA has been demonstrated through experiments, validating its performance across various application scenarios.

\end{itemize}

\section{Related Work}

\subsection{Multi-view Diffusion Models}

Diffusion models \cite{ho2020denoising,sohl2015deep,rombach2022high,peebles2023scalable} have excelled in 2D image generation, inspiring their extension to 3D content creation \cite{luo2021diffusion,karnewar2023holodiffusion,wu2024direct3d}.
However, progress in this area has been impeded by the lack of comprehensive 3D datasets. 
To overcome this challenge, recent multi-view diffusion models bypass reconstruction, directly synthesizing novel 3D views.
Watson et al. \cite{watson2022novel} pioneered the application of diffusion models to view synthesis, utilizing the ShapeNet dataset\cite{sitzmann2019scene}. 
Building on this foundation, Liu et al. introduced Zero-1-to-3 \cite{liu2023zero}, which fine-tuned a pre-trained diffusion model on a large 3D render dataset for novel view synthesis from a single RGB image.
MVDream \cite{shi2023mvdream} and ImageDream \cite{wang2023imagedream} evolved to generate consistent multi-view images from text and image prompts, respectively. A current work CAT3D \cite{gao2024cat3d} produced 3D-consistent images from various inputs. 
In contrast to the aforementioned methods, our model integrates multiple multi-view diffusion models, supporting a wider range of tasks and accommodating diverse language instructions.

\subsection{Large Multimodal Models with Vision Instruction Tuning}

Instruction tuning\cite{peng2023instruction,zhang2023instruction,ouyang2022training} is an effective method for enhancing the capability of understanding human directives of large language models like ChatGPT\cite{chatgpt} and GPT-4\cite{achiam2023gpt}. 
Recently, this concept has been extended to the visual domain, leading to the development of large multimodal models\cite{liu2024visual,zhu2023minigpt,lan2024expllm,wang2024visionllm,wang2024locllm}.
LLaVA\cite{liu2024visual} first attempted to incorporate instruction tuning into the language-image multimodal space. By constructing a high-quality dataset and fine-tuning the model, LLaVA achieves improved multimodal task performance in a wide range of multimodal tasks.
Similarly, MiniGPT-4\cite{zhu2023minigpt} fine-tuned BLIP-2\cite{li2023blip} models on synthetic multimodal instruction-following data, demonstrating enhanced reasoning and comprehension in visual-linguistic tasks.
Building on these advancements,  VisionLLM\cite{wang2024visionllm} applied visual instruction tuning to various  vision-centric tasks, 
while LocLLM\cite{wang2024locllm} focused on keypoint localization tasks, utilizing instruction tuning to improve 2D/3D keypoint detection.
Inspired by these works, we leverage visual instruction tuning to enable intelligent model selection.

\begin{figure}[t]
\centerline{\includegraphics[width=\linewidth]{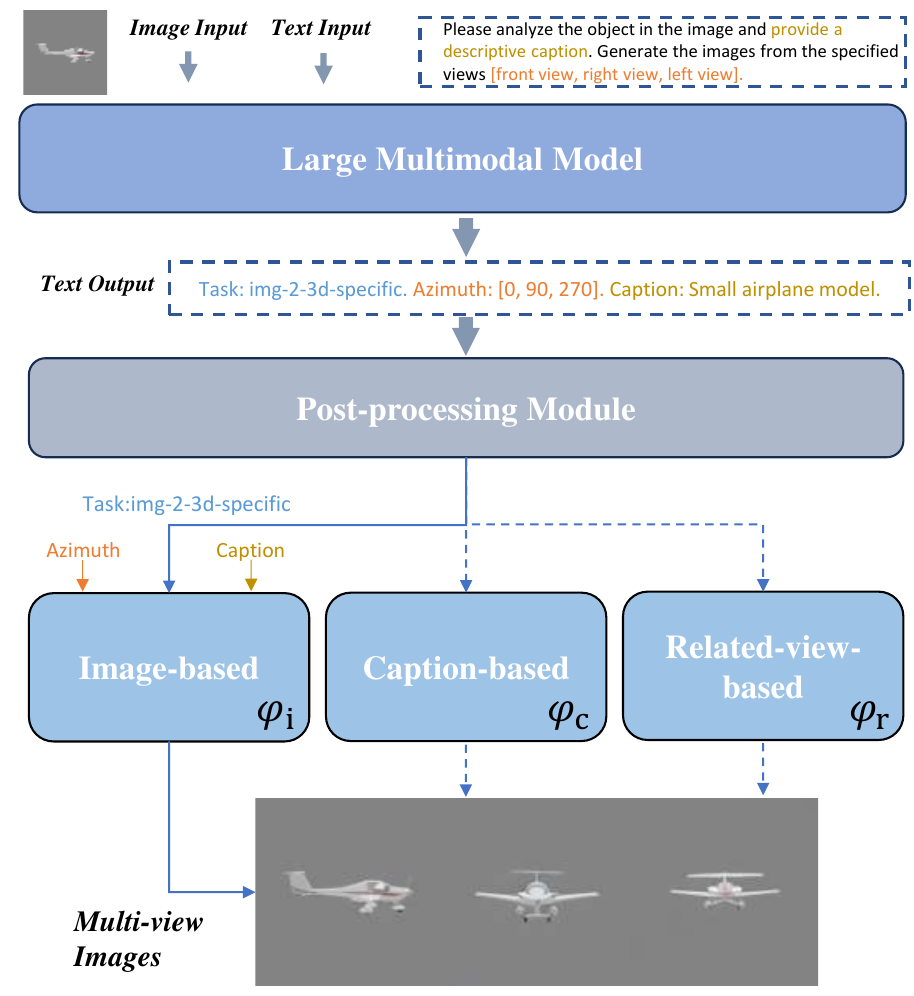}}
\caption{The overall architecture of MVLLaVA. It consists of three main components: a large multimodal model LLaVA, a post-processing module, and multi-view diffusion models. 
}
\label{overview}
\end{figure}

\section{MVLLaVA}
\subsection{Overview}
MVLLaVA aims to build an intelligent agent using a large multimodal model LLaVA to unify various novel view synthesis tasks. 
The process begins with the meticulous design of view-based instructions, specifically tailored to fine-tune LLaVA \cite{liu2024visual,liu2024improved}. As illustrated in Fig. \ref{overview},  this enhanced multimodal model is then fused with a diverse ensemble of multi-view diffusion models, fostering a seamless integration. Additionally, a post-processing module is introduced, which acts as a discerning selector, identifying the most appropriate downstream multi-view diffusion model for each task. 


\subsection{View-based Instructions}
We design five types of templates categorized into three main groups. Each group corresponds to a specific multi-view diffusion model, and we will discuss them in detail:


\subsubsection{Image-based tasks}

These tasks use a reference image to generate new images based on global viewpoint instructions, which can be separated into two categories:

\textbf{Img-2-3d-around.} This task involves generating a specified number (from 1 to 8) of images of the target object from various angles. The instruction format is: “Analyze the object in the image and provide a descriptive caption. Generate \text{\textless$n$\textgreater} images from different perspectives around the object.” where \text{\textless$n$\textgreater} is dynamically replaced.

\textbf{Img-2-3d-specific.}  In this category, users specify the desired viewpoints (e.g., left, front, right, rear) for the generated images. The instruction format includes: ``Please analyze the object in the image and provide a descriptive caption. Provide the image from the \text{\textless$v$\textgreater}." The token \text{\textless$v$\textgreater} represents any combination of the predefined viewpoints.


\subsubsection{Caption-based tasks}
In caption-based tasks, only text instructions are used as input, with caption information specified in the instructions. Similar to image-based tasks, novel views can be synthesized from either surrounding or specific viewpoints:

    \textbf{Text-2-3d-around.}       The example instruction is ``Please generate images from \text{\textless$v$\textgreater} different perspectives around the object based on the description \text{\textless$c$\textgreater}." And \text{\textless$c$\textgreater} will be replaced by the input caption.
    
    \textbf{Text-2-3d-specific.}     The input format is  ``Please provide the images from the \text{\textless$v$\textgreater} based on the description \text{\textless$c$\textgreater}."

\subsubsection{Related-view-based task} In this category, a single reference image is also used, but the instruction focuses on indicating relative changes in viewpoint rather than providing global viewpoint information.

\textbf{Img-2-3d-degree.} Its instruction format is ``please analyze the object in the image and provide a descriptive caption. Provide the image with the camera rotated by \text{\textless$d$\textgreater} degrees." Here \text{\textless$d$\textgreater} is a random rotation degree in (-360, 360).

\subsubsection{Answer format}

As shown in Table \ref{task}, we design similar templates with varying content for the five tasks. 
The task and azimuth fields are essential for all tasks, while the caption field is exclusive to the Image-based and Related-view-based tasks, which require caption generation from the reference image.
The azimuth field serves as a critical condition for multi-view diffusion models, defining global azimuth angles (e.g. 0° for a front view) that are either evenly distributed or specified in the instruction, within the range [0, 360). For related-view tasks, azimuths are simplified to the original and rotated global angles of the reference image.

\begin{table}[ht]
\centering
\caption{View-based instructions designed for novel view synthesis. The names of tasks and groups are abbreviated.}
\resizebox{1\linewidth}{!}{
\begin{tabular}{ccl}
\bottomrule
Group                          & Task   Name        & \multicolumn{1}{c}{Answer Format}                                                \\ \hline
\multirow{2}{*}{I-based}   & I-around    & Task: I-around. Azimuth: {[}d1,…,dn{]}. Caption: ...   \\
                                   & I-specific  &  Task: I-specific. Azimuth: {[}d1,…,dn{]}. Caption: ... \\ \hline
\multirow{2}{*}{C-based} & T-around   & Task: T-around. Azimuth: {[}d1,…,dn{]}.               \\
                               & T-specific & Task: T-specific. Azimuth: {[}d1,…,dn{]}.             \\ \hline
R-based             & I-degree    & Task: I-degree. Azimuth: {[}d1,d2{]}. Caption: ...     \\ \bottomrule
\label{task}
\end{tabular}}
\end{table}

\begin{figure}[tb]
\centerline{\includegraphics[width=\linewidth]{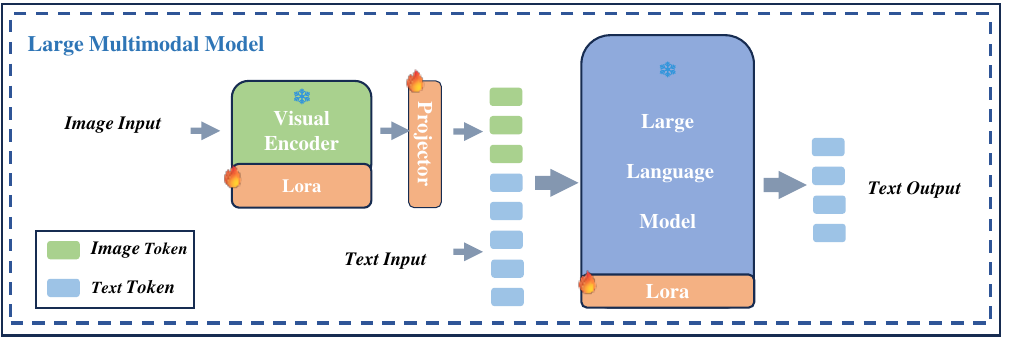}}
\caption{Training of LLaVA. We freeze the language model and the visual encoder, adding the LoRA module into both the visual encoder and the LLM.}
\label{training}
\end{figure}

\begin{figure*}[htbp]
\centerline{\includegraphics[width=0.95\linewidth]{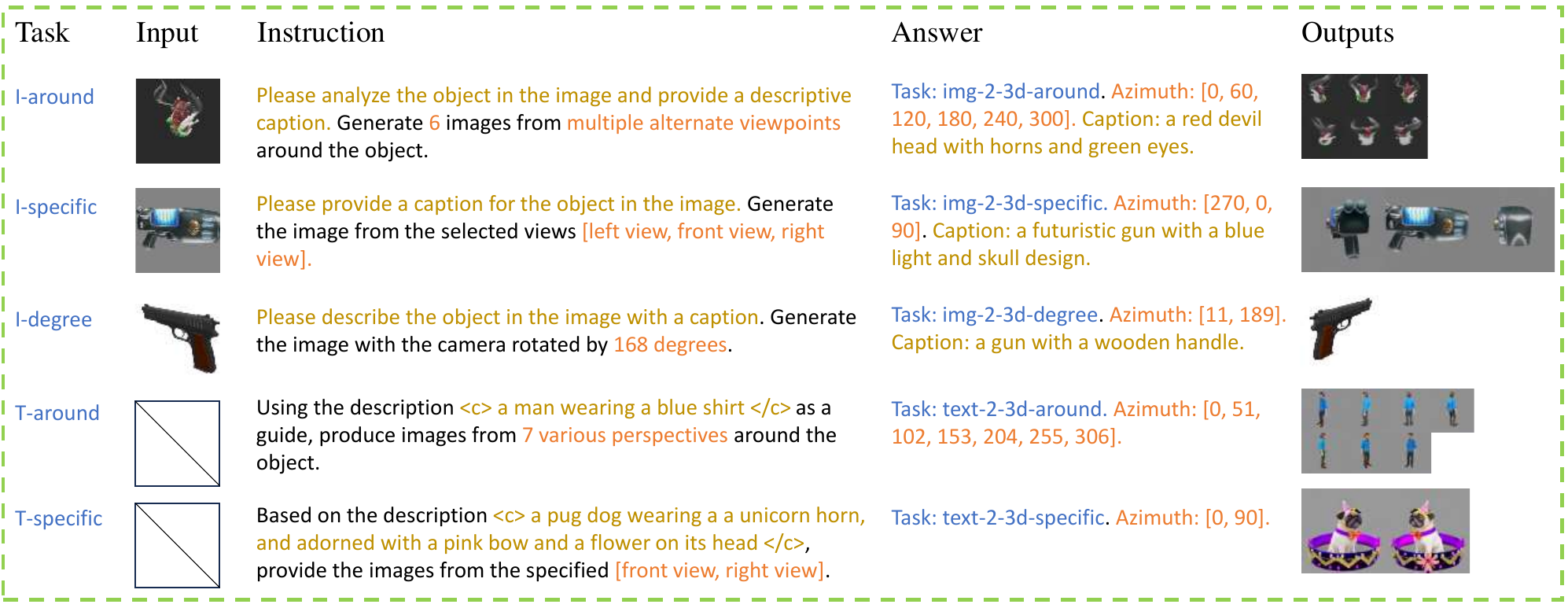}}
\caption{The Qualitative results of MVLLaVA across five different tasks, where ``I" denotes ``Img-2-3d" and ``T" denotes ``Text-2-3d".}
\label{visual_results}
\end{figure*}

\subsection{Efficient Fine-tuning}

To efficiently fine-tune LLaVA \cite{liu2024improved} with limited GPU resources and data, we freeze the language model and visual encoder while integrating Low-Rank Adaptation (LoRA) \cite{hu2021lora} into both components, as shown in Fig. \ref{training}. Specifically, LoRA is applied to the weight matrices in the attention layers, facilitating parameter-efficient optimization.

\section{Experiments}
\subsection{Implementation Details}

We use a portion of the Objaverse\cite{deitke2023objaverse} 
dataset to train and test our MVLLaVA model. 
Following the filtering approach of LGM\cite{tang2024lgm}, we remove low-quality objects and those lacking textures, resulting in a subset of 80k 3D objects. 
Subsequently, each object is then centered and rendered from 32 uniformly sampled azimuth angles, with fixed camera elevation and radius, at a resolution of 256x256.
We use Cap3D\cite{luo2024scalable} to generate captions for each object. 
When constructing the instruction dataset, we randomly select a task and a rendered image, then generate the instructions using GPT-4\cite{achiam2023gpt} based on basic prompts. 
To evaluate MVLLaVA, 1,000 instructions are randomly designed, covering five distinct tasks.


We utilize LLaVA-1.5-7B \cite{liu2024improved} as our base multimodal model, which combines the Vicuna-7B  large language model \cite{zheng2023judging} with a visual encoder (ViT) \cite{dosovitskiy2020image}. ImageDream \cite{wang2023imagedream}, MVDream \cite{shi2023mvdream}, and Zero 1-to-3 \cite{liu2023zero} are utilized as downstream models for image-based, caption-based, and related-view tasks, respectively. 
During training, the LoRA dimensions are set to 128 and are inserted into the visual encoder and LLM of LLaVA. 
The model is trained for 1 epoch, with a batch size of 2 per GPU, using an initial learning rate of 2e-4.

\textbf{Metrics.} To evaluate MVLLaVA's performance, four distinct metrics are designed to assess the model's capabilities:

\begin{itemize}
\item Task Accuracy (TA): Measures the model's accuracy in distinguishing task types. A prediction is considered correct if the predicted task fully matches the actual task.

\item Azimuth Accuracy (AA): For the img-2-3d-degree task, this metric verifies the consistency of azimuth changes before and after rotation. For other tasks, it evaluates the accuracy of predicted azimuths against the actual values.


\item Caption BLUE-score (CB): The CB metric evaluates the quality of predicted captions using the BLEU score \cite{papineni2002bleu}, which provides an assessment of word accuracy. 

\item Caption CLIP-score (CC): The CC metric assesses the semantic similarity between generated captions and actual captions by CLIP\cite{radford2021learning} score similarity. 
    
\end{itemize}

\subsection{Ablation Study}

We conduct an ablation study by fine-tuning different components of LLaVA, resulting in three models: MVLLaVA-L (fine-tuning only the LLM), MVLLaVA-V (fine-tuning only the ViT), and MVLLaVA (fine-tuning both the ViT and LLM). 
As shown in Table \ref{ablation}, MVLLaVA consistently outperforms the others, underscoring the significance of joint fine-tuning. Specifically, MVLLaVA-L and MVLLaVA exhibit similar performance levels, while MVLLaVA-V performs notably worse, primarily due to the lack of ViT fine-tuning for caption-based tasks. This highlights the necessity of fine-tuning both components to achieve optimal performance across diverse tasks.

\begin{table}[ht]
\centering
\caption{The ablation study of MVLLaVA.}
\begin{tabular}{cccccc}
\bottomrule
Model    & TA $\uparrow$     & AA $\uparrow$      & CB $\uparrow$ &CC $\uparrow$ &Avg. $\uparrow$  \\ \hline
MVLLaVA-L   & 1.0  & 0.969 & 0.557 &0.797 & 0.830 \\
MVLLaVA-V & 0.6 & 0.49 & 0.548  &  0.780 & 0.604 \\
MVLLaVA   & \textbf{1.0} & \textbf{0.975}  & \textbf{0.566} & \textbf{0.802}  &\textbf{0.836}\\ 
\bottomrule
\label{ablation}
\end{tabular}
\end{table}

\begin{table}[t]
\centering
\caption{The quantitative results of MVLLaVA across five different tasks, where ``I" denotes ``Img-2-3d" and ``T" denotes ``Text-2-3d".}
\begin{tabular}{cccccc}
\bottomrule
\diagbox{Tasks}{Metrics}        & TA      & AA  & CB & CC & \\ \hline
I-around   & 1.0 & 1.0   & 0.570  & 0.805 \\
I-specific & 1.0 & 1.0   & 0.566  & 0.801 \\
T-around   & 1.0 & 1.0 & -  & -  \\
T-specific & 1.0 & 1.0 & -  & - \\
I-degree   & 1.0 & 0.875   &  0.563 &0.799  \\ \hline
Avg. & 1.0 & 0.975  & 0.566 & 0.802 \\ \bottomrule
\label{acc}
\end{tabular}
\end{table}

\subsection{Evaluation on Different Tasks}

We evaluate MVLLaVA and present the experimental results across different tasks in Table \ref{acc}. MVLLaVA achieves a task accuracy of 100\% and an azimuth accuracy of 97.5\%, demonstrating its proficiency in recognizing novel view synthesis tasks and accurately interpreting azimuth angles. 
Additionally, the CB score of 0.566 and CC score of 0.802 highlight its capabilities in caption prediction, supporting downstream image generation models.

\begin{table}[ht]
\centering
\caption{The quantitative results of MVLLaVA compared with SOTA models.}
\begin{tabular}{cccccc}
\bottomrule
Model      & TA $\uparrow$     & AA $\uparrow$      & CB $\uparrow$ &CC $\uparrow$ &Avg.  $\uparrow$ \\ \hline
GPT-4o\cite{gpt4o} & 0.989 & \textbf{0.993}  &  0.497 & 0.753 & 0.807\\
Claude-3.5-Sonnet\cite{claude35sonnet} &0.993 &0.977  & 0.475 & 0.745 &0.797\\ 
MVLLaVA   & \textbf{1.0} & 0.975  & \textbf{0.566} & \textbf{0.802}  &\textbf{0.836}   \\ 
\bottomrule
\label{sota}
\end{tabular}
\end{table}

\subsection{Comparisons with State-of-the-art Models}

We compare our MVLLaVA with state-of-the-art large multimodal models, GPT-4o\cite{gpt4o} and Claude-3.5-Sonnet\cite{claude35sonnet}. 
As shown in Table \ref{sota}, MVLLaVA outperforms the current SOTA models across all metrics except for the AA metric. Moreover, MVLLaVA achieves an overall average metric score of 0.836, surpassing GPT-4o's score of 0.807. Notably, our model utilizes a 7B version of LLaVA, which highlights its strong performance despite having fewer parameters compared to the SOTA models. Additionally, while GPT-4o and Claude-3.5-Sonnet require supplementary information such as task descriptions and answer formats, MVLLaVA only needs predefined instructions, offering users greater convenience.

\subsection{Qualitative Results}
\begin{figure*}[htbp]
\centerline{\includegraphics[width=0.95\linewidth]{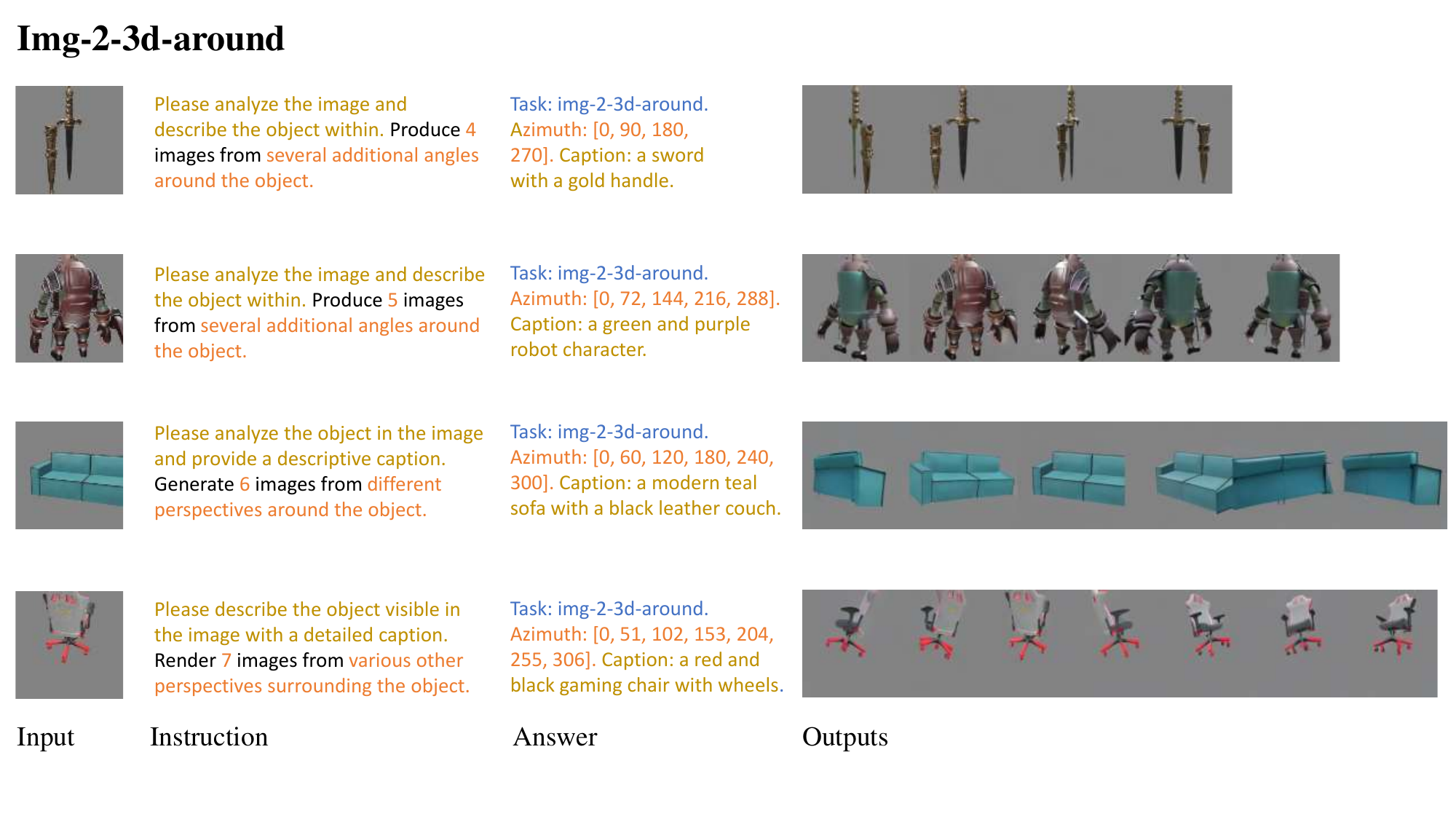}}
\caption{The Qualitative results of MVLLaVA on the Img-2-3d-around task. ImageDream\cite{wang2023imagedream} is used for multi-view diffusion.}
\label{1}
\end{figure*}

\begin{figure*}[htbp]
\centerline{\includegraphics[width=0.95\linewidth]{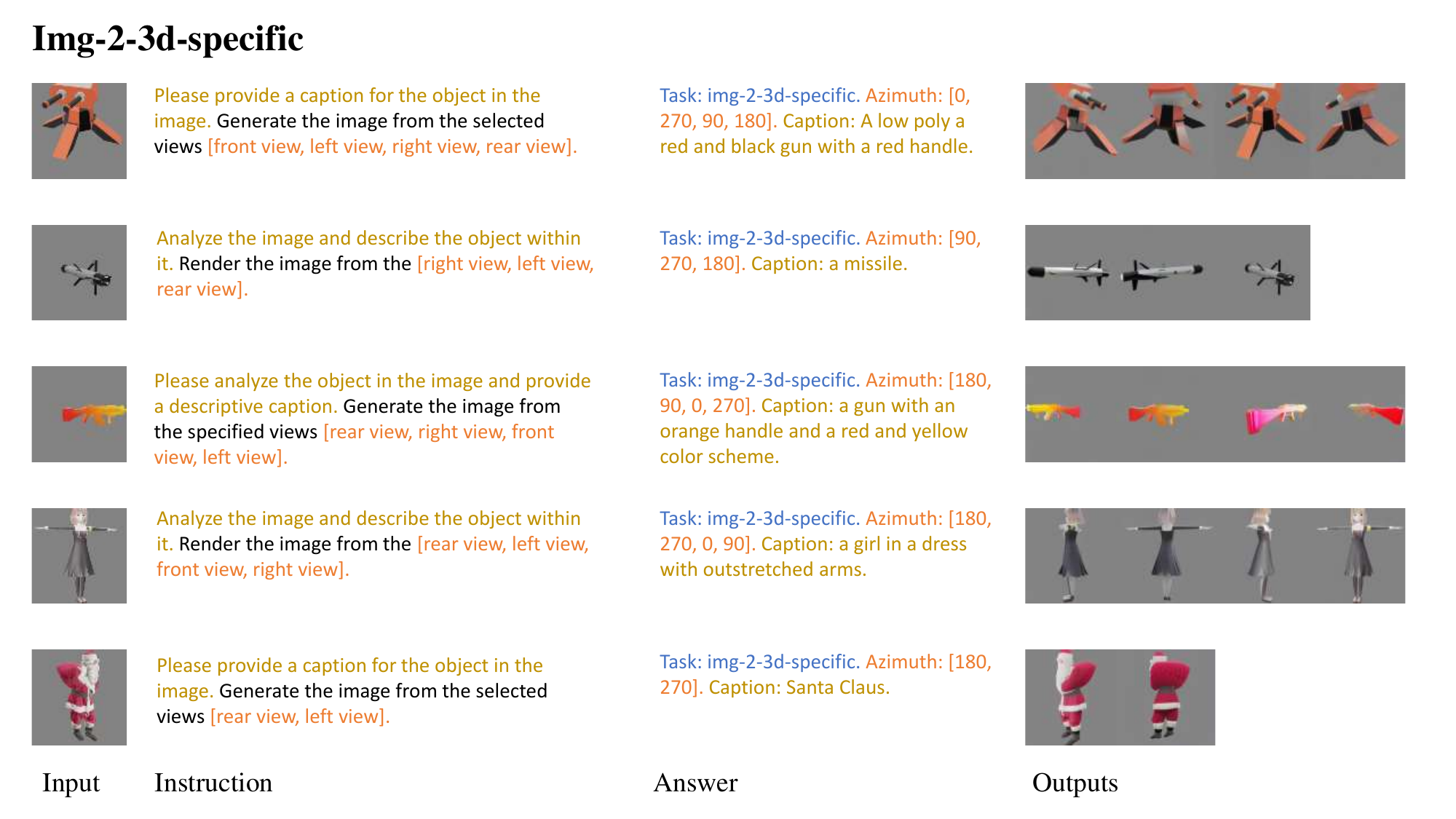}}
\caption{The Qualitative results of MVLLaVA on the Img-2-3d-specific task. ImageDream\cite{wang2023imagedream} is used for multi-view diffusion.}
\label{2}
\end{figure*}

\begin{figure*}[htbp]
\centerline{\includegraphics[width=0.95\linewidth]{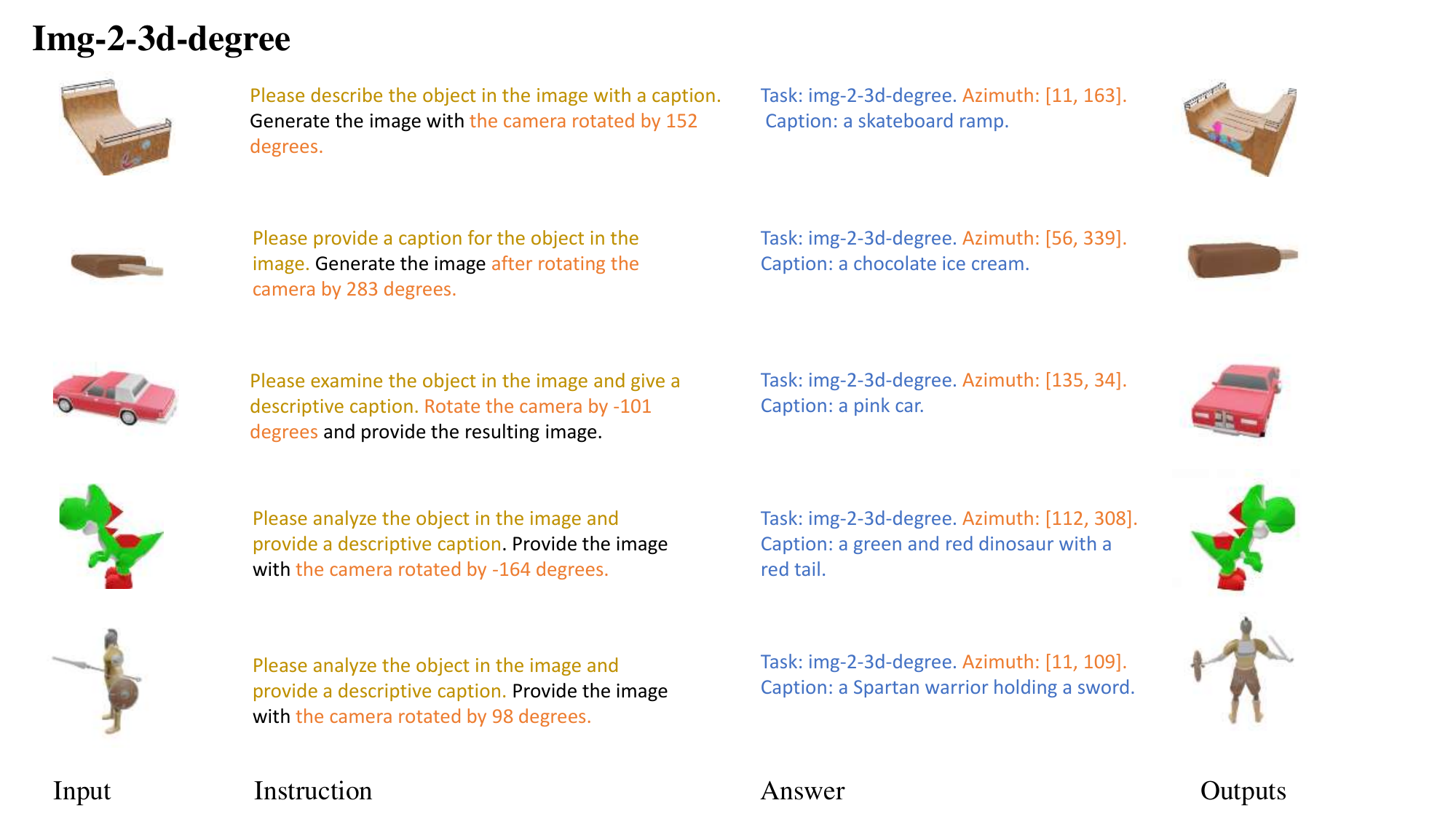}}
\caption{The Qualitative results of MVLLaVA on the Img-2-3d-degree task. Zero123\cite{liu2023zero} is used for multi-view diffusion.}
\label{3}
\end{figure*}

\begin{figure*}[htbp]
\centerline{\includegraphics[width=0.95\linewidth]{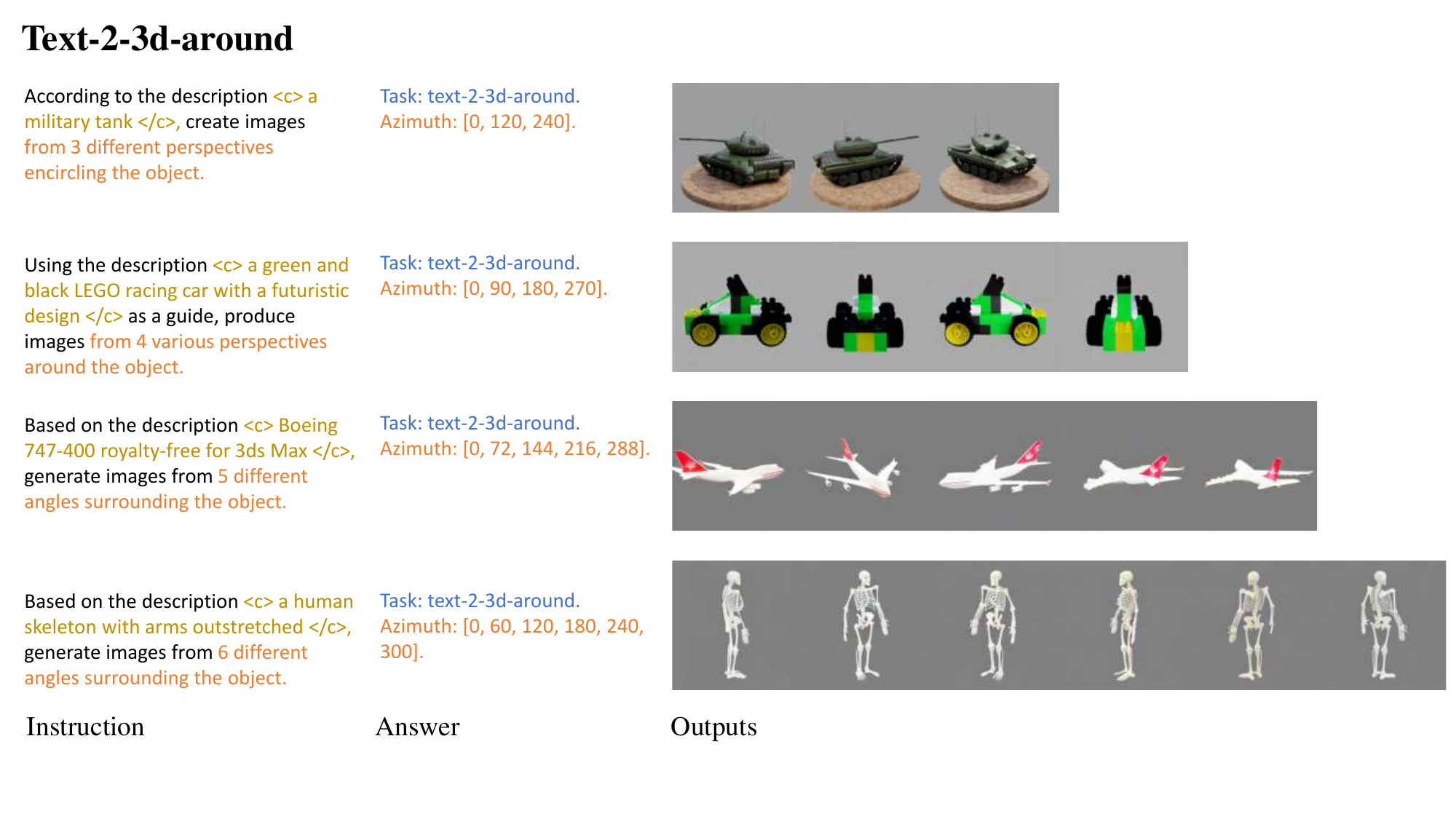}}
\caption{The Qualitative results of MVLLaVA on the text-2-3d-around task. MVdream\cite{shi2023mvdream} is used for multi-view diffusion.}
\label{4}
\end{figure*}

\begin{figure*}[htbp]
\centerline{\includegraphics[width=0.95\linewidth]{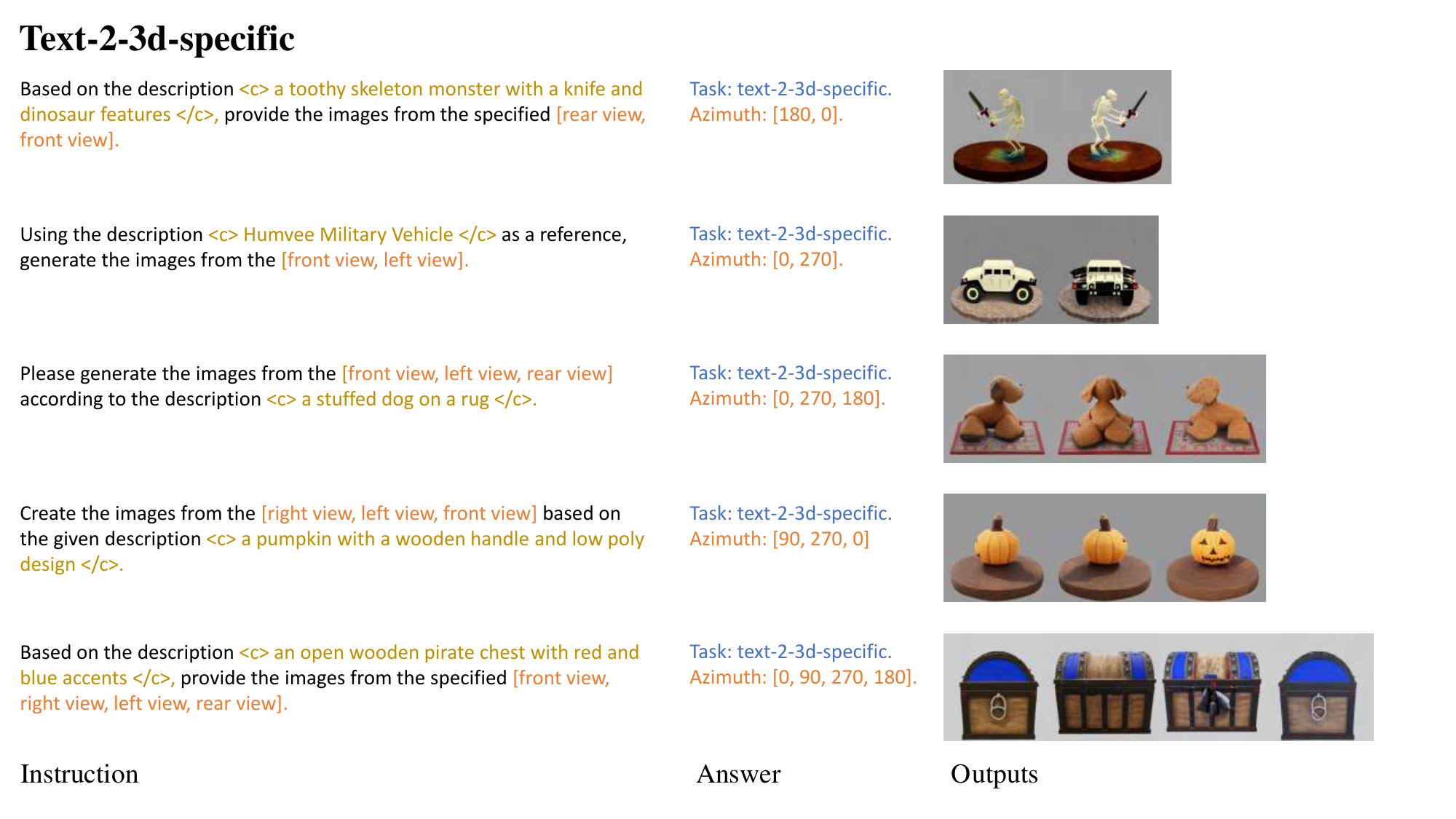}}
\caption{The Qualitative results of MVLLaVA on the text-2-3d-specific task. MVdream\cite{shi2023mvdream} is used for multi-view diffusion.}
\label{5}
\end{figure*}

Fig. \ref{visual_results} visualizes MVLLaVA's multi-view generation outputs based on various task instructions. 
MVLLaVA effectively identifies the relevant tasks and accurately extracts the required information for downstream models, producing the desired outputs. The flexibility in input instructions further highlights its versatility in handling diverse human instructions.  More results are visualized from Fig. \ref{1} to Fig. \ref{5}.
It is noteworthy that MVLLaVA features a loosely coupled design between the multimodal large model and the multi-view diffusion models. This design allows MVLLaVA to be easily applied to various multi-view diffusion models with minimal adjustments to the post-processing module, specifically involving slight modifications in the handling of azimuth angles.




\section{Conclusion}

In this paper, we introduce MVLLaVA, an intelligent agent designed to overcome the limitations of current multi-view diffusion models in novel view synthesis. By integrating multiple diffusion models with LLaVA, MVLLaVA offers exceptional flexibility, adapting to a wide range of inputs and instructions. The careful design and efficient fine-tuning of task-specific instruction templates further enhance its performance across various scenarios. Experimental results validate the robustness and versatility of MVLLaVA, highlighting its potential as a powerful solution for unified multi-view generation tasks.

\newpage



\bibliographystyle{IEEEtran}
\bibliography{IEEEexample}

\end{document}